\documentclass{bmvc2k}

\usepackage{booktabs}
\usepackage{array}
\usepackage{diagbox}
\usepackage{caption}
\usepackage{algorithm}
\usepackage{algorithmic}
\usepackage{enumitem}
\newcolumntype{C}[1]{>{\centering\arraybackslash}p{#1}}
\title{Using Synthetic Corruptions to Measure Robustness to Natural Distribution Shifts}

\addauthor{Alfred LAUGROS}{alfred.laugros@atos.net}{12}
\addauthor{Alice CAPLIER}{alice.caplier@grenoble-inp.fr}{1}
\addauthor{Matthieu OSPICI}{matthieu.ospici@atos.net}{2}

\addinstitution{
 Universite Grenoble Alples,\\
 France\\
 \ \\
}
\addinstitution{
 Atos,\\
 France\\
}

\runninghead{Student, Prof, Collaborator}{BMVC Author Guidelines}

\begin{document}

\maketitle

\begin{abstract}
Synthetic corruptions gathered into a benchmark are frequently used to measure neural network robustness to distribution shifts. However, robustness to synthetic corruption benchmarks is not always predictive of robustness to distribution shifts encountered in real-world applications. In this paper, we propose a methodology to build synthetic corruption benchmarks that make robustness estimations more correlated with robustness to real-world distribution shifts. Using the overlapping criterion, we split synthetic corruptions into categories that help to better understand neural network robustness. Based on these categories, we identify  three relevant parameters to take into account when constructing a corruption benchmark that are the (1) number of represented categories, (2) their relative balance in terms of size and, (3) the size of the considered benchmark. In doing so, we build new synthetic corruption selections that are more predictive of robustness to natural corruptions than existing synthetic corruption benchmarks.
\end{abstract}

\section{Introduction}
Neural networks have been shown to be sensitive to distribution shifts such as common corruptions \cite{inet_c}, adversarial examples \cite{ref_adv} or background changes \cite{terra_inco}. When deployed in a production context, neural networks often encounter samples that come from potentially drastically different distributions between train and test time. Because of this, they obtain lower performances in practical applications compared to the performances observed on their test sets.  During model conception and training, the exhaustive variety of input distributions is very rarely accessible, once deployed in a production scenario. Consequently, it is necessary to make neural networks more robust to distribution shifts.

Some methods have been proposed to make neural networks more robust to distribution shifts \cite{ant,adv_prop,noisy_student}. To estimate if these methods are useful in practice, we need to establish benchmarks that measure robustness to distribution shifts. Traditionally used approaches consist in measuring performances of models on out-of-distribution samples, i.e. samples that come from a different distribution than the one used to get the training samples. The underlying idea is that the better the performance of a model on an unseen distribution, the better one can expect to be robust by potential distribution shifts.

But, there is no guarantee that the robustness measured using one particular distribution transfers to other distributions: a model robust to colorimetry variations is not necessarily robust to background changes. To address this issue, we generally use several distributions during the testing phase, to create more diverse out-of-distribution test samples. We assume that the more a model is robust to a large diversity of distribution shifts, the more this model is likely to generalize to other unseen distributions. For this reason, finding new distributions to draw more diverse test samples, can be useful to improve robustness estimations.

Various distribution shifts can be obtained by using synthetic corruptions such as Gaussian noise, rotations, contrast loss... The robustness of a model can be estimated by testing its performances on a test set that has been corrupted using various image transformations. In this paper, we make a distinction between synthetic and natural corruptions. Synthetic corruptions correspond to modeled images transformations that are used to corrupt images such as translations or quantizations. On the other hand, natural corruptions are distribution shifts arising naturally in real world applications \cite{taori_nat}. In this study, we do not consider transformations especially crafted to fool neural networks such as adversarial attacks \cite{ref_adv,evasion}.

Constituting a benchmark of naturally corrupted samples is costly. It requires to draw samples from a distribution that is not covered by existing datasets, and to label the gathered samples. Samples corrupted with synthetic corruptions are cheaper to gather. They can be obtained by corrupting already labeled images. Then, robustness to synthetic corruption benchmarks is often used as a proxy for robustness to natural corruptions \cite{cure_tsr,stylized_imagenet,segmentation_rob}. In some contexts, this approach seems relevant, for example, synthetic blur robustness is highly predictive of the robustness to real-world blurs \cite{many_face_rob}. However, other experiments show that robustness to traditionally used synthetic corruptions is not correlated with robustness to natural corruptions \cite{taori_nat}. Consequently, we do not really know in which circumstances synthetic corruption robustness can be used as a proxy for natural corruptions. In this paper, we address this issue by making the following contributions:

\begin{itemize}[itemsep=0pt]
	\item{We show that some corruption selections are much more predictive of robustness to natural corruptions than others. }
	\item{We propose three high-level parameters that help to determine which corruption selections are more correlated in terms of robustness with natural corruptions. Specifically, given an initial set of synthetic corruptions, we split this set into categories. These categories are built such as the corruptions belonging to the same category overlap (they are correlated in terms of robustness), while the corruptions belonging to different categories do not. Based on these categories, we identify 3 parameters to take into account while building a synthetic corruption benchmark: (1) the number of represented corruption categories (2) the balance among categories (3) the size of benchmarks.}
	\item{We present a methodology that takes into account these parameters to generate benchmarks, and we use it to get corruption selections that are more predictive of robustness to natural corruptions than existing synthetic corruption benchmarks.}
\end{itemize}

\section{Related Works \label{sec:sota}}

\textbf{Natural Corruption Benchmarks.} Several natural corruption benchmarks have been proposed to estimate robustness of image classifiers to distribution shifts. For instance, SVSF is a store front classification dataset that reveals the natural corruptions that arise when varying three parameters: camera, year and country \cite{many_face_rob}. The SI-SCORE dataset focuses on the robustness to other parameters such as object size, location and orientation \cite{si_score}. Robustness to background changes is also a widely studied topic \cite{terra_inco}. Several robustness benchmarks have been built to measure robustness of ImageNet classifiers. ImageNet-A is a challenging benchmark, constructed by selecting images that are misclassified by various ResNet-50 architecture based models \cite{inet_a}. ImageNet-Sketch \cite{inet_sketch} is an alternative ImageNet validation set containing hand-drawn skectches. ImageNet-V2 \cite{inet_v2} has been built by replicating the ImageNet construction process. Because of some statistical biases in the image selection \cite{bias_inet_v2}, a distribution shift is observed between ImageNet and ImageNet-V2. ObjectNet \cite{onet} is a set of images that contains objects that have been randomly rotated or taken with various backgrounds and viewpoints. ImageNet-R contains artistic renditions of ImageNet object classes \cite{many_face_rob}. ImageNet-D has been recently proposed to provide additional challenging distribution shifts (quickdraw, infograph...) \cite{inet_d}.

\textbf{Using synthetic corruptions to measure robustness to natural distribution shifts.} Natural corruption benchmarks are costly to constitute, so synthetic corruption benchmarks are often used as a proxy for estimating robustness in various computer vision tasks such as face recognition \cite{face_rec_noise}, object detection \cite{cc_object_detection}, image segmentation \cite{segmentation_rob}, saliency region detection \cite{cc_gaze},  traffic sign recognition \cite{cure_tsr} and scene classification \cite{scene_classif}. ImageNet-C \cite{inet_c} is used to estimate robustness of ImageNet classifiers, it contains fifteen corruptions which can be classified into noises, blurs, weather and digital corruptions. Inspired from psychophysics, RichardWebster et al. proposed to estimate robustness using sequences that contain corrupted images derived from a single image \cite{psyphy}. The corruption amount  progressively changes throughout each sequence. Similarly, ImageNet-P \cite{inet_c} contains sequences of subtly corrupted images and measures the probability of flipping predictions ($mFR$ metric) between two successive sequence images.

Although these benchmarks are widely used to estimate image classifier robustness \cite{noisy_student,augmix,stylized_imagenet,adv_prop}, Taori et al. question the idea of using synthetic corruptions as a proxy for natural corruptions \cite{taori_nat}. Indeed, they show for instance that the robustness to ImageNet-C is not predictive of the robustness to some natural corruption benchmarks such as ImageNet-V2. On the other hand, Hendrycks et al. give some examples of natural corruption benchmarks that do correlate in terms of robustness with the ImageNet-C corruptions \cite{inet_c}. Then, the circumstances under which synthetic corruption robustness is predictive of real-world robustness are not clearly defined. In this paper, we address this issue by presenting attributes of synthetic corruption benchmarks that largely influence the way robustness to these benchmarks transfer to natural corruptions.

\section{Background \label{sec:back}}

\textbf{Corruption Overlappings.} Our benchmark generation methodology is based on the notion of corruption overlapping. Two synthetic corruptions overlap when they are correlated in terms of robustness. For instance, it has been demonstrated that corruptions that damage high frequencies in images (noises, blurs...) overlap \cite{fourier,laugros19}. It has been shown that a benchmark should not contain a couple of corruptions $c_1,c_2$ such as $c_1$ overlaps much more with the other corruptions of the benchmark than $c_2$ \cite{overlap_score}. Otherwise, the considered benchmark gives too much importance to the robustness towards some kinds of corruptions compared to others. The overlapping score metric \cite{overlap_score} has been recently proposed to measure to what extent two corruptions $c_1$ and $c_2$ overlap:
\begin{center}
$ O_{c_1,c_2} = \max\{0,\frac{1}{2}*\left(\frac{R^{m1}_{c2}-R^{standard}_{c2}}{R^{m2}_{c2}-R^{standard}_{c2}}   +     \frac{R^{m2}_{c1}-R^{standard}_{c1}}{R^{m1}_{c1}-R^{standard}_{c1}}\right)\} \ \ \ \ \ \ \ \ \ \ (1)$
\end{center}
$m_1$, $m_2$ and $standard$ are models with the same architecture. $m1$ and $m2$ have been respectively trained with data augmentation on $c_1$ and $c_2$; $standard$ is only trained on clean samples. $R^m_c$ is the ratio between the accuracy of $m$ on samples corrupted with $c$ and the accuracy of $m$ on not-corrupted samples. The idea behind the overlapping score is that the more a data augmentation with $c_1$ makes a model robust to $c_2$ and conversely, and the more we can suppose that $c_1$ and $c_2$ are correlated in terms of robustness. The overlapping range value is $[0\mbox{-}1]$. The higher this score is, the more the considered corruptions overlap.

\textbf{Robustness Metrics.} In all experiments, we measure the robustness of an image classifier $f$ to a distribution $P$ by computing the residual robustness: $R(f,P) = A_{i.i.d}(f)-A_P(f)$. $A_{i.i.d}(f)$ and $A_P(f)$ are the accuracies of $f$ respectively computed with $i.i.d.$ samples (independent and identically distributed samples with regard to the training set of $f$) and samples drawn from $P$. Other robustness metrics could have been used \cite{taori_nat, inet_c}, but we choose the residual robustness because it is how robustness is generally estimated in industrial applications: it is the accuracy drop caused by a distribution shift. We note that comparing the residual robustness of two models to a distribution shift, requires to check that the accuracies on $i.i.d.$ samples of the two models are comparable. This condition is verified in all our experiments. In this paper, the robustness of a model $f$ to a synthetic corruption benchmark $bench$, refers to the mean of the residual robustnesses of $f$ computed with the corruptions of $bench$. 

\textbf{Experimental Set-up.} All overlapping scores computed in this paper, are obtained using the ImageNet-100 dataset (a subset of ImageNet that contains every tenth ImageNet class by WordNetID order \cite{wordnet_id}), the ResNet-18 architecture, and exactly the same training hyperparameters as the ones used in the paper introducing the overlapping score \cite{overlap_score}. 

In some experiments, we evaluate correlations in terms of robustness between benchmarks. To do this, we use a set of models that cover various neural network architectures, sizes and training methodologies. The idea is to obtain a set of models representative of the diversity of image classifiers that can be used. Some of the selected models have been trained with data augmentation on either adversarial examples \cite{spatial_rob,ant,pgd_aug} or synthetic corruptions \cite{fastautoaugment,stylized_imagenet,augmix,many_face_rob,cutmix}. Some others have trained leveraging a large amount of unlabeled data using self-supervised learning \cite{mopro,noisy_student,wsl,ssl}. We also select models having non-standard architectures \cite{adv_prop,anti_alias}. The gathered models are displayed in Table \ref{tab:rob_models}. We also use the plain counterparts (trained without any robustness intervention) of these models.

\begin{table}
\begin{center}
\begin{small}
\begin{tabular}{C{91mm}|C{28mm}}
\toprule
Models Trained with a Robustness Intervention & Plain Counterpart\\
\midrule

FastAutoAugment \cite{fastautoaugment}; Worst-of-10 spatial data augmentation using the following transformation space: $\pm 3$ pixels $\pm 30$ degrees \cite{spatial_rob} & ResNet18  \\
\midrule
ANT\textsuperscript{3x3} \cite{ant}; SIN Augmentation \cite{stylized_imagenet}; Augmix \cite{augmix}; DeepAugment \cite{many_face_rob}; MoPro \cite{mopro}; RSC \cite{rsc}; Adversarial Training: $L_{inf}, \epsilon = 4/255$ \cite{pgd_aug} & ResNet-50\\
\midrule
Noisy Student Training \cite{noisy_student}; AdvProp \cite{adv_prop} & EfficientNet-0\\
\midrule
Anti-Aliased \cite{anti_alias} & DenseNet-121 \\
\midrule
Cutmix \cite{cutmix} & ResNet-152\\
\midrule
Weakly Supervised Pretraining \cite{wsl}; Semi-Supervised Pretraining \cite{ssl} & ResNeXt-101-32x16d \\
\bottomrule
\end{tabular}
\end{small}
\end{center}
\caption{Selected models trained with a robustness intervention, we also use their plain counterpart.\label{tab:rob_models}}
\end{table}
\section{Corruption Categories\label{sec:cat}}

\begin{figure}
\begin{center}
\includegraphics[scale=0.114]{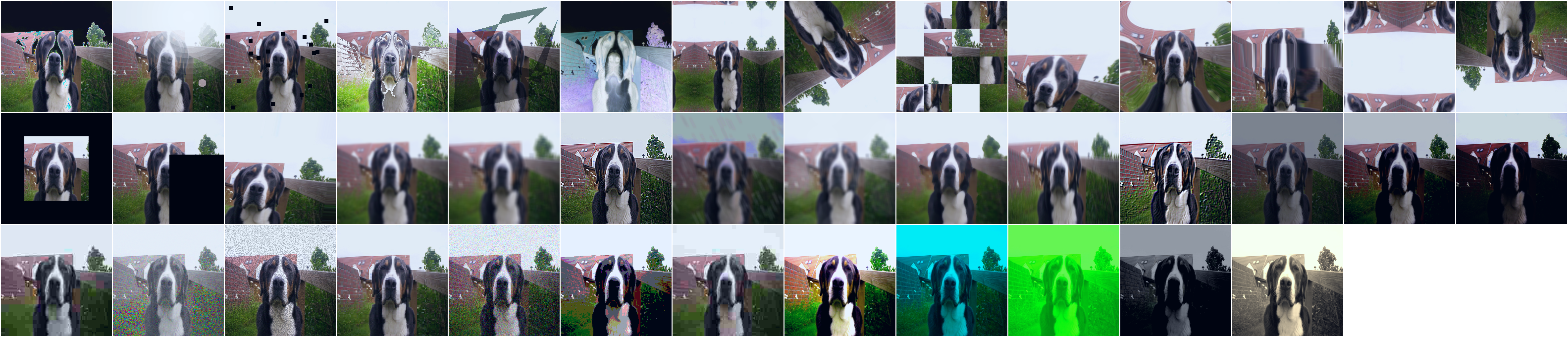}
\end{center}
   \caption{Candidate corruptions displayed in the same order as the corruption names of Figure \ref{fig:overlaping_scores_matrix}. \label{fig:cc_mosaic}}
\label{fig:short}
\end{figure}

\begin{figure}
\begin{center}
\includegraphics[width=0.98\textwidth]{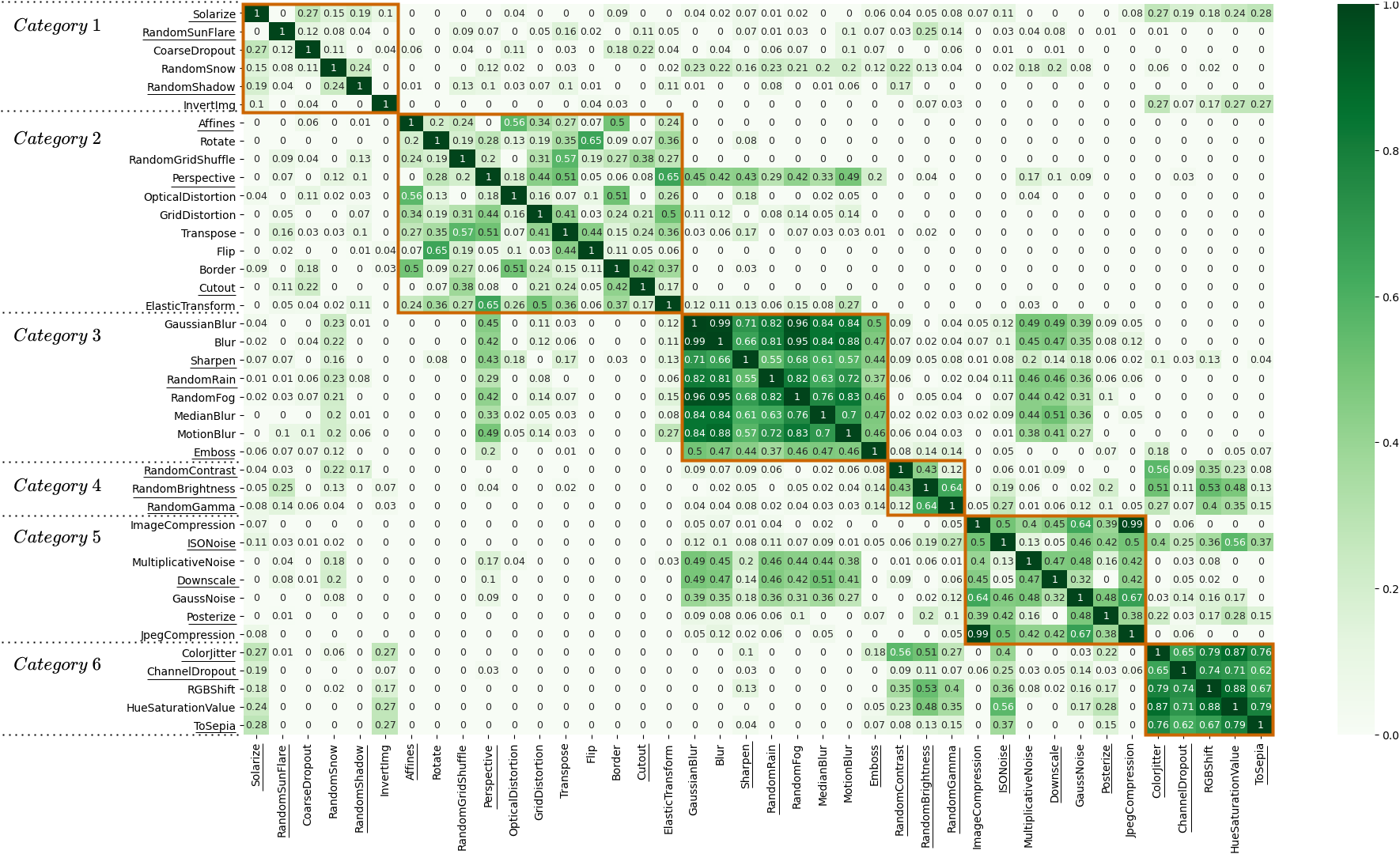}
\end{center}
   \caption{Overlapping scores computed using all the possible candidate corruption couples. The scores computed with SCC are in the 6 orange squares: one square for each of the 6 categories. \label{fig:overlaping_scores_matrix}}
\label{fig:short}
\end{figure}

There are a lot of possible synthetic corruptions that can be included in a benchmark. Constructing a corruption benchmark requires to pick some corruptions among all possible candidates. Here, we consider a list of 40 candidates whose names can be seen in the abscissa of Figure \ref{fig:overlaping_scores_matrix} and they are illustrated in Figure \ref{fig:cc_mosaic}. An other list of corruptions could have been selected, but most of the corruptions that are usually included in existing benchmarks \cite{cure_tsr,face_rec_noise,inet_c} can be found in these candidates: blurs, noises, contrast loss... The corruptions are implemented using the albumentations library \cite{albumentations}, implementation details can be found at  \url{https://github.com/bds-ailab/common_corruption_benchmark}.

The 40 considered corruptions form a heterogeneous set. It is difficult to determine the number and the kinds of corruptions to be included in a robustness benchmark a priori. In this paper, we propose a method to select groups of corruptions that make robustness estimations more correlated with robustness to natural corruptions. 

The first step of our method is to compute the overlapping scores between candidate corruptions: here we use the corruptions displayed in Figure \ref{fig:cc_mosaic}. Each corruption $c$ is then associated with a vector that contains all the overlapping scores computed using $c$ and any other corruptions. The second step is to split the candidate corruptions into categories, such as the overlapping score vectors of the corruptions belonging to the same category are correlated; while the overlapping score vectors of the corruptions belonging to different categories are not. To achieve it, we cluster our candidates using their associated 40-dimensional vector of overlapping scores. We use the K-means algorithm increasing progressively the number of centroids $K$. We note that increasing $K$, raises on average the correlations between the overlapping vectors of the Same Category Corruptions (SCC), which is consistent with our goal; but it also raises the correlations between the vectors of Different Category Corruptions (DCC), which is not desired. We choose to stop increasing $K$ at $K=6$, when the mean of all the Pearson correlation coefficients computed using SCC overlapping vectors becomes higher than 0.5. The obtained categories can be seen in Figure \ref{fig:overlaping_scores_matrix}.

We observe that all these categories do not contain the same number of corruptions. We also notice that SCC can be associated with a human visual perception interpretation for most of the categories. Indeed, $Category\ 2$ to $6$ in Figure \ref{fig:overlaping_scores_matrix}, could be respectively called \textit{spatial transformations, blurs, lightning condition variations, fine-grain artifacts and color distortions}. Note that the SCC of $Category\ 1$ overlap way less than the ones of other categories because it contains more heterogeneous corruptions. Corruptions of $Category\ 1$ would likely have been distributed between more refined categories by using additional corruptions in our initial set of candidates.

\textbf{Empirical Evaluation.} We conduct an additional experiment to verify the relevance of the built corruption categories. For each corruption $c$ among the candidate corruptions displayed in Figure \ref{fig:cc_mosaic}, we compute the residual robustness of the twenty-one models displayed in Table \ref{tab:rob_models} with the ImageNet validation set corrupted with $c$. Each candidate corruption $c$ is now associated with a vector that contains the twenty-one robustness scores computed using $c$. For each possible couple of candidate corruptions, we compute the Pearson correlation coefficient using the two robustness score vectors associated with the corruptions of the considered couple. The mean correlation obtained using SCC is 0.67, while the one obtained using DCC is 0.10. This experiment confirms the relevance of the built corruption categories: SCC are in practice correlated in terms of robustness while DCC are not.

\section{Synthetic Corruption Selection Criteria\label{sec:exp}}
	
We introduce the definition of some terms used in this paper. The size of a benchmark is the number of corruptions this benchmark contains. Each time a benchmark $bench$, contains a corruption $c$ that belongs to the corruption category $CC$, we say that $CC$ is represented in $bench$; and $c$ is called a representative of $CC$ in $bench$. In this section, we identify three parameters of synthetic corruption benchmarks that influence the way robustness to these benchmarks is correlated with robustness to natural corruptions. These parameters are: (1) the number of corruption categories represented (2) the balance among categories (3) the size of benchmarks. We make an ablation study in each of the three following sections to demonstrate the importance of each parameter.

\subsection{Number of Corruption Categories Represented in Benchmarks \label{sec:param1}}

Each corruption category displayed in Figure \ref{fig:overlaping_scores_matrix} contains image transformations that modify different attributes in images. For instance, $Category\ 6$ contains essentially corruptions that modify colorimetry; while $Category\ 4$ contains corruptions that modify contrast and brightness. As a consequence, the features modified in one category, are mostly different from the ones modified in the other categories. Then, we make the assumption that the more corruption categories are represented in a benchmark, the more this benchmark takes into account a large diversity of attribute modifications. Distribution shifts due to natural corruptions generally change a lot of attributes at the same time: background, resolution, viewpoint... Then, intuitively, the largest the number of represented categories in a benchmark is, the more this benchmark is likely to make robustness estimations predictive of robustness to natural corruptions. To verify this intuition, we propose to use Algorithm \ref{alg:bench_gene} to build several benchmarks that have various numbers of represented categories.

\begin{algorithm} 
\caption{Corruption Benchmark Generation Algorithm}
\begin{small}
\label{alg:bench_gene}
\begin{algorithmic}
\REQUIRE A group of candidate corruptions split into several corruption categories $cat$
\REQUIRE $n$: the number of categories represented in the returned benchmark
\REQUIRE $k$: the number of representatives of the categories represented in the returned benchmark
\STATE Randomly select $n$ categories in $cat$
\STATE For each selected category, randomly select $k$ distinct corruptions of this category
\RETURN A benchmark that contains the selected corruptions
\end{algorithmic}
\end{small}
\end{algorithm}

Using the corruption categories displayed in Figure \ref{fig:overlaping_scores_matrix}, we run Algorithm \ref{alg:bench_gene} for several $n,k$ couples: (2,3),(3,2),(6,1),(4,3),(6,2),(6,3). We repeat this process until we obtain a group of 1000 different benchmarks for each of the considered $n,k$ couples. We note that a benchmark generated using $n=4,k=3$ contains 3 representatives of 4 out of 6 categories. We want to measure if increasing the number of represented categories $n$ makes robustness estimations of benchmarks more correlated with robustness to natural corruptions. To verify this, we propose the Algorithm \ref{alg:syn_nat_corr}, that measures to what extent the robustness estimations made by a group of synthetic corruption benchmarks, are on average correlated with the robustness to one natural corruption benchmark. For each of the benchmark groups generated using Algorithm \ref{alg:bench_gene}, we run Algorithm \ref{alg:syn_nat_corr} for each of the following natural corruption benchmarks: ImageNet-A \cite{inet_a}, ImageNet-R \cite{many_face_rob}, ImageNet-V2 \cite{inet_v2}, ImageNet-Sketch \cite{inet_sketch} and ObjectNet \cite{onet}. The group of neural networks used to run Algorithm \ref{alg:syn_nat_corr} is the set of models presented in Table \ref{tab:rob_models}. The obtained scores are displayed in Table \ref{tab:param1} ($n,k$ columns). We compute the mean p-value associated with each of these scores: they are all lower than $0.02$, i.e., these correlations are statistically significant.

\begin{algorithm} 
\caption{Estimates the average correlation between the robustness to one natural corruption benchmark and the robustness to synthetic corruption benchmarks}
\begin{small}
\label{alg:syn_nat_corr}
\begin{algorithmic}
\REQUIRE $SB$ a group of synthetic corruption benchmarks
\REQUIRE $TNN$ a group of trained neural networks
\REQUIRE $NB$ a benchmark of naturally corrupted samples
\STATE $scores_1 \gets$ the residual robustness of all models in $TNN$ computed with $NB$
\STATE For each benchmark $sb$ in $SB$:
\STATE\hspace{\algorithmicindent} $scores_2 \gets$ the residual robustness of all models in $TNN$ computed with $sb$
\STATE\hspace{\algorithmicindent} Get the Pearson correlation coefficient between $score_1$ and $score_2$
\RETURN the mean of the correlation coefficients computed in the loop
\end{algorithmic}
\end{small}
\end{algorithm}

\begin{table}
\begin{center}
\begin{small}
\begin{tabular}{p{21mm}||C{7mm}C{7mm}C{7mm}|C{7mm}C{7mm}|C{7mm}||C{7mm}C{7mm}C{7mm}}
\toprule
\diagbox[width=24mm,height=9mm]{Natural}{Synthetic} & $n,k$ $2,3$ &  $n,k$ $3,2$ &  $n,k$ $6,1$ &  $n,k$ $4,3$ & $n,k$  $6,2$ & $n,k$ $6,3$ & INet-C & INet-P & INet-S2N\\
\midrule
INet-A & 0.667 &	0.680 &	0.701 &	0.707 &	0.721 &	0.729 & 0.642 & 0.561 & 0.726\\
INet-R & 0.635 &	0.653 &	0.678 &	0.682 &	0.697 &	0.707 & 0.565  & 0.437 & 0.691\\
INet-V2  & 0.691 &	0.733 &	0.751 &	0.757 &	0.777 &	0.785 & 0.857 & 0.920 & 0.794\\
ObjectNet  &  0.695 &	0.732 &	0.757 &	0.763 &	0.789 &	0.796 & 0.807 & 0.823 & 0.814\\
INet-S  &  	0.651 &	0.674 &	0.695 &	0.701 &	0.717 &	0.725 & 0.641 & 0.511 & 0.713\\
\midrule
Mean & 0.668 & 0.694 & 0.716 & 0.720 & 0.740 & \textbf{0.748} & 0.702 & 0.650 & \textbf{0.748}\\
\bottomrule
\end{tabular}
\end{small}
\end{center}
\caption{Correlation scores computed using Algorithm \ref{alg:syn_nat_corr} between natural corruption benchmarks and synthetic corruption benchmarks. The $n,k$ couples define the bencharks generated using  Algorithm \ref{alg:bench_gene}.\label{tab:param1}}
\end{table}

The higher the score of a group of synthetic corruption benchmarks of Table \ref{tab:param1} is, the more the considered group makes on average robustness estimations correlated with the robustness to the natural corruption benchmark used to compute this score.  To only study the effect of the number of represented categories $n$ in benchmarks, we only compare the scores of Table \ref{tab:param1} obtained using benchmarks that have the same size. So, we compare the benchmarks of 6 corruptions generated using the $(n,k)$ couples $(2,3), (3,2)$ and $(6,1)$. We see that the obtained scores increase with $n$ for all the tested natural corruption benchmarks. Similarly, for the benchmarks of 12 corruptions generated using the $(n,k)$ couples $(4,3)$ and $(6,2)$, the obtained scores are higher for $n=6$ than $n=4$. This experiment confirms the idea that increasing the number of categories represented in synthetic corruption benchmarks, makes robustness to these benchmarks more predictive of robustness to natural corruptions.

\subsection{Balance Among Categories}

We consider that the balance among categories represented in a benchmark is preserved, when all represented categories of this benchmark have the same number of representatives. For instance, the balance among categories of a benchmark $bench$ that contains the Gaussian noise, iso-noise, multiplicative noise and color-jitter corruptions is not preserved: $bench$ contains three representatives of $Category\ 5$ and one representative of $Category\ 6$ (see Figure \ref{fig:overlaping_scores_matrix}). Obviously $bench$ is biased towards texture damaging robustness rather than colorimetry variation robustness. Intuitively, the robustness to a benchmark biased towards a few kinds of feature modifications, is not likely to be predictive of robustness to natural corruptions that change a large diversity of features in images. Consequently, preserving the balance among categories, should help to build benchmarks that make robustness estimations more correlated with robustness to natural corruptions.

We conduct an experiment to verify this intuition. We note $std$, the standard deviation computed using all the numbers of representatives of the categories represented in a benchmark. The $std$ of benchmarks generated using Algorithm \ref{alg:bench_gene} are null: their balance among category is preserved. We propose to get new benchmarks that have higher $std$ by using the substitution operation. A substitution randomly removes a corruption $c_1$ from a benchmark $bench$ and adds to it a corruption $c_2$ randomly selected in the set of candidates. But, $c_1$ and $c_2$ are selected such as three conditions are respected: (1) the represented categories of $bench$ do not change (2) $std$ of $bench$ strictly increases (3) $c_2$ is not already in $bench$.
 
We consider $group$: a set of 1000 benchmarks that have been generated using Algorithm \ref{alg:bench_gene} with $n=6,k=2$. We get 5000 new benchmarks, by substituting from 1 to 5 corruptions of each benchmark in $group$. The obtained benchmarks have various $std$: (0.6, 0.8, 1.0, 1.2, 1.4, 1.5, 1.8, 2.2). We group together all the benchmarks with the same $std$, the obtained groups contain all more than 200 benchmarks. For each of these groups, we run Algorithm \ref{alg:syn_nat_corr} for each of the following natural corruption benchmarks: ImageNet-A, ImageNet-R, ImageNet-V2, ImageNet-Sketch and ObjectNet. The group of neural networks used to run Algorithm \ref{alg:syn_nat_corr} is the set of models presented in Table \ref{tab:rob_models}. The obtained results are displayed in Table \ref{tab:param2}. We also compute the mean p-value associated with each score of this table, they are all lower than $0.02$. We observe that the scores of Table \ref{tab:param2} decrease as the $std$ of corruption benchmarks increases. We repeat the experiment carried out in this section, using different benchmarks generated with Algorithm \ref{alg:syn_nat_corr} with $n=5,k=3$ and $n=6,k=3$. For both $n,k$ couples, the measured mean correlations also diminish as $std$ of benchmarks increases. These experiments confirm that benchmarks for which balance among represented category is preserved, make robustness estimations more correlated with robustness to natural corruptions.

\begin{table}
\begin{center}
\begin{small}
\begin{tabular}{p{18mm}||C{6mm}C{6mm}C{6mm}C{6mm}C{6mm}C{6mm}C{6mm}C{6mm}C{6mm}}
\toprule
{} & 0.0&  0.6&  0.8&  1.0 & 1.2&  1.4&  1.5&  1.8&  2.2 \\
\midrule
ImageNet-A & 0.721 &	0.709 	&0.711 	&0.712 	&0.710 	&0.695 &	0.691 	&0.685 	& 0.685 \\
ImageNet-R & 0.697 & 	0.698 	&0.694 	&0.682 	&0.689 	&0.676 	&0.662 &	0.659 &	0.658\\
ImageNet-V2  & 0.777 &	0.777 	&0.782 	&0.771& 	0.767 &	0.749 &	0.750 &	0.710 &	0.687 \\
ObjectNet  & 0.789 & 0.798 & 0.781 & 0.786 & 0.777 &	0.761 &	0.744 &	0.713 &	0.698 \\
ImageNet-S & 0.717 	&0.708 &	0.708& 	0.715 &	0.703 &	0.699 &	0.688 &	0.675 &	0.673 \\
\midrule
Mean & 0.740 & 0.738 & 0.735 & 0.733 & 0.729 & 0.716 & 0.707 & 0.688 & 0.680 \\
\bottomrule
\end{tabular}
\end{small}
\end{center}
\caption{Correlations computed with Algorithm \ref{alg:syn_nat_corr} using natural corruption benchmarks (lines) and groups of synthetic corruption benchmarks that have various $std$ values (columns). \label{tab:param2}}
\end{table}

\subsection{Corruption Benchmark Size} 

In Table \ref{tab:param1}, we observe that the scores obtained with the benchmarks generated using the $n,k$ couples $(6,1)$, $(6,2)$ and $(6,3)$; increase with $k$. We notice that rising $k$ for a fixed $n$ when using Algorithm \ref{alg:bench_gene}, is equivalent to increase the size of generated benchmarks while conserving the balance among categories and the number of represented categories. Then, it appears that increasing the size of synthetic corruption benchmarks also helps to make robustness estimations more correlated with natural corruptions. To explain this, we see in Figure \ref{fig:overlaping_scores_matrix} that SCC are not completely correlated in terms of robustness. In other words, the representatives of the same category do not make exactly the same feature modifications. So, having more representatives in each category makes benchmarks measure the robustness to a larger range of feature modifications. Since natural corruptions modify a wide diversity of features in images, having more representatives per category (increasing $k$ for a fixed $n$) should make robustness to corruption benchmarks more predictive of robustness to natural corruptions, which can explain the obtained results.

\subsection{Comparison with Existing Synthetic Corruption Benchmarks}

We want to set the three parameters identified in the previous sections to get the corruption selections that are the most correlated as possible in terms of robustness with natural corruptions. In others words, we want to build the largest benchmarks, that represent all the categories displayed in Figure \ref{fig:overlaping_scores_matrix} and have their balance among categories preserved. These benchmarks can be obtained by running the Algorithm \ref{alg:bench_gene}, with $n=6$ and $k$ the largest as possible. In our case, the largest possible $k$ is three because the smallest category displayed in Figure \ref{fig:overlaping_scores_matrix} contains only three corruptions. 

Now, we want to determine if the benchmarks generated using $n=6$, $k=3$ are more predictive of robustness to natural corruptions than two existing synthetic corruption benchmarks : ImageNet-C and ImageNet-P \cite{inet_c}. In the same way as in Section \ref{sec:param1}, we run Algorithm \ref{alg:syn_nat_corr} using the models displayed in Table \ref{tab:rob_models} with ImageNet-C, to estimate its correlation in terms of robustness with several natural corruption benchmarks. We repeat this process with ImageNet-P, but since this benchmark is not meant to be used with the residual robustness metric, we use its associated metric $mFR$ \cite{inet_c} instead to measure robustness towards this benchmark. The results are displayed in Table \ref{tab:param1}. The p-values associated with the scores computed using ImageNet-P and ImageNet-C are all lower than 0.05.

We observe that the benchmarks generated using $n=6$, $k=3$ are on average much more correlated in terms of robustness with ImageNet-A, ImageNet-Sketch and ImageNet-R than ImageNet-C and ImageNet-P. The contrary is observed for ImageNet-V2. For ObjectNet, the obtained scores are relatively close. The last line of Table \ref{tab:param1} shows that the benchmarks generated using our methodology are on average more predictive of robustness to natural corruptions than ImageNet-C and ImageNet-P. It would be interesting to identify the reasons why the results obtained with ImageNet-V2 contrast with the general tendency.

For convenience, we provide an example of corruption selection picked from the benchmarks obtained using $n=6;k=3$ that we call ImageNet-Syn2Nat. It corresponds to the generated benchmark with the SCC that overlap the least which each other. The idea is to avoid getting a benchmark that contains corruptions that are almost equivalent. Obviously, other ways to pick a single benchmark could be used and we intend to investigate the best ones in further works. The correlations with natural corruption benchmarks of ImageNet-Syn2Nat are displayed in Table \ref{tab:param1}, and the names of its corruptions are underlined in Figure \ref{fig:overlaping_scores_matrix}.

\section{Conclusion}
We proposed a method to split a set of synthetic corruptions into some categories using the overlapping score. We showed that such categories are useful to better understand and address robustness of neural networks. By using corruption categories, we identified three parameters that are important to consider while building a corruption benchmark: the number of categories represented, the balance among categories and the size. Taking into account these parameters helps to build corruption benchmarks that make robustness estimations more correlated with robustness to natural corruptions. We hope that these works will help to better understand in which circumstances robustness to synthetic corruptions transfers to natural corruptions.

Future works include using a larger set of candidate corruptions and trying other clustering strategies. We would also like to build more refined corruption categories than the ones presented in Figure \ref{fig:overlaping_scores_matrix}. Besides, it could be interesting to consider additional benchmarks in the study such as ImageNet-D \cite{inet_d}. Most importantly, it would be valuable to apply the presented methodology to computer vision tasks where natural corruption benchmarks are particularly costly to build such as image segmentation or 3D vision.

\bibliography{egbib}
\end{document}